\documentclass{sigchi-ext}
\usepackage[T1]{fontenc}
\usepackage{textcomp}
\usepackage[scaled=.92]{helvet} % for proper fonts
\usepackage{graphicx} % for EPS use the graphics package instead
\usepackage{balance}  % for useful for balancing the last columns
\usepackage{booktabs} % for pretty table rules
\usepackage{ccicons}  % for Creative Commons citation icons
\usepackage{ragged2e} % for tighter hyphenation
\usepackage[normalem]{ulem}

\def\plaintitle{Beyond One-Size-Fits-All: Adapting Counterfactual Explanations to User Objectives} 
\def\emptyauthor{}
\def\plainkeywords{Counterfactual; Explanations; XAI; Adversarial; Human-Centered Explanations}

\title{Beyond One-Size-Fits-All: Adapting Counterfactual Explanations to User Objectives}

\numberofauthors{3}
\author{%
  \alignauthor{%
    \textbf{Orfeas Menis Mastromichalakis}\\
    \affaddr{Artificial Intelligence and Learning Systems Laboratory}\\
    \affaddr{School of Electrical and Computer Engineering}\\
    \affaddr{National Technical University of Athens} \\
    \email{menorf@ails.ece.ntua.gr} }  \alignauthor{%
    \textbf{Jason Liartis}\\
    \affaddr{Artificial Intelligence and Learning Systems Laboratory}\\
    \affaddr{School of Electrical and Computer Engineering}\\
    \affaddr{National Technical University of Athens}\\
    \email{jliartis@ails.ece.ntua.gr} } \vfil \alignauthor{%
    \textbf{Giorgos Stamou}\\
    \affaddr{Artificial Intelligence and Learning Systems Laboratory}\\
    \affaddr{School of Electrical and Computer Engineering}\\
    \affaddr{National Technical University of Athens}\\
    \email{gstam@cs.ntua.gr} } }

\definecolor{linkColor}{RGB}{6,125,233}
\hypersetup{%
  pdftitle={\plaintitle},
  pdfauthor={\emptyauthor},
  pdfkeywords={\plainkeywords},
  bookmarksnumbered,
  pdfstartview={FitH},
  colorlinks,
  citecolor=black,
  filecolor=black,
  linkcolor=black,
  urlcolor=linkColor,
  breaklinks=true,
}

\begin{document}

%% For the camera ready, use the commands provided by the ACM in the Permission Release Form.
\CopyrightYear{2024}
\setcopyright{rightsretained}
\conferenceinfo{CHI'24}{May  11--16, 2024, Honolulu, HI, USA}
\isbn{ HCXAI Workshop}
\doi{ }
%% Then override the default copyright message with the \acmcopyright command.
\copyrightinfo{\acmcopyright}

\maketitle

% Uncomment to disable hyphenation (not recommended)
% https://twitter.com/anjirokhan/status/546046683331973120
\RaggedRight{} 

% Do not change the page size or page settings.
\begin{abstract}
Explainable Artificial Intelligence (XAI) has emerged as a critical area of research aimed at enhancing the transparency and interpretability of AI systems. Counterfactual Explanations (CFEs) offer valuable insights into the decision-making processes of machine learning algorithms by exploring alternative scenarios where certain factors differ. Despite the growing popularity of CFEs in the XAI community, existing literature often overlooks the diverse needs and objectives of users across different applications and domains, leading to a lack of tailored explanations that adequately address the different use cases. In this paper, we advocate for a nuanced understanding of CFEs, recognizing the variability in desired properties based on user objectives and target applications. We identify three primary user objectives and explore the desired characteristics of CFEs in each case. By addressing these differences, we aim to design more effective and tailored explanations that meet the specific needs of users, thereby enhancing collaboration with AI systems.  % Abstracts should be about 150 words and are required.
\end{abstract}


% ACM Classfication


% \ccsdesc[300]{Human-centered computing~Haptic devices}
% \ccsdesc[100]{Human-centered computing~User studies}

% Print the classification codes
% Please use the 2012 Classifiers and see this link to embed them in the text: \url{https://dl.acm.org/ccs/ccs_flat.cfm}

\section{Introduction}

% 1-2 sentences about XAI and its improtance. 
Explainable Artificial Intelligence (XAI) has become increasingly indispensable as AI systems are integrated into various facets of society. It aims to elucidate the opaque decision-making processes of machine learning algorithms, offering transparency and useful insights, and fostering trust among multiple stakeholders from AI engineers to end-users. 
Counterfactual Explanations \cite{wachter2017counterfactual}, stemming from decades of studies in philosophy \cite{lewis2013counterfactuals} and psychology on counterfactual thinking \cite{byrne2016counterfactual}, delve into why a specific outcome occurred by considering alternative scenarios where certain factors or events were different.
In the realm of artificial intelligence, these explanations provide invaluable insights into how alterations in input features would have influenced an AI system's output, thereby facilitating a deeper understanding of its behavior and decision-making mechanisms. Due to their contrastive nature, it has been claimed \cite{MILLER20191} that CFEs are close to how humans perceive explanations, therefore gaining popularity among the XAI community. 
% By elucidating the causal relationships between input features and model predictions, counterfactual explanations contribute to the interpretability and trustworthiness of AI systems, making them essential tools for both developers and end-users.

% Problem statement / motivation 
While numerous existing works delve into the desired characteristics of counterfactual explanations \cite{verma2020counterfactual, karimi2020survey, guidotti2022counterfactual, verma2021counterfactual}, they often approach them with a unified strategy encompassing multiple objectives including detecting biases, providing actionable recourse, increasing trust, and enhancing understandability. However, this approach overlooks the fact that the diverse needs and objectives of users across various applications and domains necessitate different properties for counterfactual explanations. Consequently, the explanations generated may fail to adequately address all use cases, as the requirements for one user's target could directly conflict with another's.
This phenomenon of conflicting motivations and ambiguity of objectives in XAI has been thoroughly discussed in literature \cite{MythosInterpretability, CynthiaStop}.
Our work aligns with numerous works that call for contextualized design, development, and evaluation of explanations in terms of application \cite{norkute2021ai}, target audience \cite{sokol2020one, liartis2023searching, dervakos2022computing, dhanorkar2021needs}, and end-goal \cite{wolf2019explainability, guidotti2022counterfactual}.

% Therefore, there is a pressing need to distinguish between the distinct use cases in which counterfactual explanations are applied. By doing so, we can pinpoint the specific desired characteristics for each use case, rather than attempting to create counterfactuals that aim to cover all scenarios, which is inherently infeasible.

% What we do 
In this position paper, we advocate for a more nuanced understanding of counterfactual explanations, recognizing that the desired properties of these explanations can vary significantly depending on the user's objectives and target applications.
Building upon the use cases discussed in existing literature \cite{karimi2020survey, guidotti2022counterfactual} and our prior experience on counterfactual explanations \cite{chooseIJCAI} and their evaluation \cite{filandrianos-etal-2023-counterfactuals} we identify three main user objectives, and explore the desired properties of the counterfactual explanations in each case. By acknowledging these differences, we can design and develop more tailored and effective explanations that address the specific needs of users across a range of scenarios, enabling them to become better collaborators with AI systems.

% if needed we can add a small section about the theory of counterfactual explanations like "background" but I think this might not be necessary if we include it somehow in the intro. 

\section{Preliminaries}
In this section, we will explain some concepts that we use throughout this paper.
We assume that there is an AI system under examination that accepts an input $x$ and produces an output $y$.
For example, the AI system might be an automated loan approval system deployed by a bank. In this case, $x$ would be a client's application that might include information such as age, gender, occupation, income, number of existing loans, etc. and $y$ would be the AI system's decision, approve or reject.
A \emph{Counterfactual Explanation} (CFE) is a new input $x'$ that produces a different output $y' \neq y$.
The new input $x'$ is referred to as the \emph{counterfactual instance}.
The system that produces CFEs, may be referred to as the \emph{counterfactual explainer}.
% $x'$ is also required to be close to $x$ by some metric. The metric used is not standard and differs among counterfactual explainers.
The differences between $x$ and $x'$ are also referred to as the \emph{counterfactual edit}.
The counterfactual edit explains to the user what changes they need to make to the input to receive a different output.
In the automated loan approval example, a counterfactual edit may suggest to a user with a rejected application, that they need to change occupation or pay off one of their existing loans for their application to be approved.

Some counterfactual edits might be impossible to implement.
For example, it would be impossible for a person to decrease their age.
Counterfactual edits that only affect mutable features are referred to as \emph{actionable} \cite{guidotti2022counterfactual}.
Actionability in some cases is objective (e.g. one cannot change their race, or decrease their age), but in other cases it can be subjective and differ among users (e.g. although for some it might be possible to change their occupation, for others it might be very hard, or even infeasible). It is therefore desirable for an explainer to treat the actionability of the input features according to user preferences.

Some counterfactual instances are less common than others or even unattainable in the real world.
For example, it is very rare for a person to have an income of \$1 million or higher while working as a teacher, and no one can have an age of -1.
Counterfactual instances that are coherent with a reference population are referred to as \emph{plausible} or \emph{feasible} \cite{poyiadzi2020face, guidotti2022counterfactual}.
Plausible instances are also said to be close to the data manifold \cite{verma2021counterfactual}, or in-distribution, while implausible instances are said to be far from the data manifold and out-of-distribution (OOD).

In academic literature, there is a rich discussion of various counterfactual properties that influence their efficacy and their intended applications.
For instance, counterfactual edits that suggest the modification of only a few inputs are referred to as \emph{sparse} \cite{verma2021counterfactual}, which is often a desired property since it makes the edit clear and concise.
% This is often a desired property since it makes the edit clear and concise.
% It may be easier for a user to focus on a few goals of moderate difficulty, rather than a long list of goals even if they are individually easier.
% This is a property that some users may find more desirable than others and the level of sparsity which is necessary to keep the edit clear is subjective.
However, in this work, we mainly focus on actionability and plausibility, since they are the main differentiating factors among the three user objectives discussed in the subsequent section. 

\section{User Objectives}
In this section, we delve into the analysis of three distinct use cases of counterfactual explanations, focusing on the properties of actionability and plausibility.
% While most existing literature on counterfactual explanations aligns with the primary desired characteristics, we focus specifically on two properties commonly cited across these works. The first, termed ``actionability'', underscores the importance of considering the mutability and actionability of features when suggesting changes within counterfactual explanations. For instance, a counterfactual explanation should refrain from altering immutable features like one's birthplace or race. The second property, referred to by various names including ``adherence to data manifold'', ``plausibility'', and ``feasibility'', pertains to ensuring that the resultant data point of the counterfactual explanation remains within the distribution (or data manifold) of a reference population.
Throughout our examination of the three diverse use cases --namely, when an end-user seeks advice to achieve a desired outcome, when a user aims to investigate an AI system's behavior, detecting biases, flaws, or inconsistencies, and finally when a user endeavors to identify vulnerabilities within the system and fortify it against attacks-- we demonstrate that these properties are not universally desired. Instead, depending on the end task, one or both of these properties may impose limitations on the task and offer diminished insights to the user.
% The main principle is common, but some user objectives/use cases have specific restrictions that should be applied to CFEs to be useful. Therefore a modular system could support all 3 use cases that we discuss here by allowing to enable and disable restrictions. 

\subsection{Outcome Fulfillment}
In this use case, the end-user is seeking advice on how to modify the input to the AI system to achieve a different output.
In the example of the automated loan approval system, the end-user would be a bank customer whose loan application was rejected and seeks advice on how to adjust it to secure approval.
This is the most restrictive use case, which requires both actionable and plausible counterfactuals.
Actionability guarantees that the user is asked to only modify input features that can change, while plausibility guarantees that the user is asked to modify them in ways that are reasonable.
% Counterfactual explanations must generate data points within the data manifold and provide edits that are easy for the user to implement.
The focus is on providing practical guidance for achieving desired outcomes while ensuring the actionability of suggested edits.
% It is often stated that sparse edits, that suggest changes in a small subset of the input variables, are preferable to edits that suggest small changes to many of the input variables.
% This might not always be the case for the end user, so some modularity that controls the desired sparsity level should be provided.
% Many systems that produce counterfactual explanations use a predefined level of sparsity even though the underlying algorithm allows for such modularity.

\subsection{System Investigation}
In this use case, the user seeks to understand the behavior of the AI system within the domain of in-distribution data.
Counterfactual explanations are employed to uncover potential biases in the model or to reveal inconsistencies.
For instance, an AI engineer developing a loan approval system may investigate whether the model exhibits biases related to sensitive attributes like gender or race. Additionally, the engineer may assess whether the model's decision-making aligns with human intuition and logical reasoning. For example, a counterfactual edit proposing a reduction in income to secure loan approval would run contrary to intuitive expectations, revealing a flaw in the system's behavior.
This is a less restrictive use case since any input feature can be present in the counterfactual edit, even non-actionable ones. If we restrict the edits to be actionable, we will fail to detect potential biases on immutable features such as race or gender. However, this investigation concerns the real-life application of the system, therefore plausibility is a desired property of the counterfactual instances.
% The emphasis here is on detecting biases within the data manifold, even if the suggested edits are not necessarily actionable.
Presenting a multitude of counterfactuals is very important in this use case, since it is through their contrast that biases are revealed.
E.g. one counterfactual edit might be an increase of income by \$10,000 and another might be an increase of income by \$1,000 and a change of gender. This looser income requirement for a different gender would reveal a gender bias.
% Sparsity is also highly desirable in this use case because it makes biases easier to single out.

\subsection{Vulnerability Detection}
In this use case, the user aims to identify potential weaknesses or vulnerabilities in the AI system.
Counterfactual explanations are used to test the robustness of the model to small perturbations or out-of-distribution inputs.
For example, a security engineer may want to ensure that slight changes to input data, or inconsistencies like leaving fields empty or providing invalid values, do not compromise the integrity of the system.
The primary emphasis here lies on robustness, where considerations of plausibility and actionability pose potential conflicts with the user's objectives as they could impede the detection of vulnerabilities to attacks involving random noise or out-of-distribution permutations.
% Sparsity is sometimes undesirable in this use case since many tiny changes can be much harder for humans to spot than one big change (especially in very high-dimensional domains such as images) and therefore make the AI system more vulnerable to attacks.
Imperceptible changes to the input that produce a different output, often referred to as adversarial examples \cite{goodfellow2014explaining} also fall under this category. Although they are usually treated as a separate concept and some counterfactual explainers explicitly suppress them, technically they are a specific use case of counterfactual edits. It would be preferable to specify whether a counterfactual explainer fulfills this use case, rather than treat it as an entirely different concept.

\section{Conclusions}
In this paper, we have underscored the importance of a nuanced understanding of counterfactual explanations in the realm of Explainable Artificial Intelligence. By recognizing the variability in desired properties based on user objectives and target applications, we have advocated for a tailored approach to the design and development of CFEs. Our analysis of three main user objectives has revealed that the desired characteristics of CFEs differ significantly depending on the end task, highlighting the necessity of considering user needs in the explanation process. Through this study, we have demonstrated the limitations of a one-size-fits-all approach to CFEs and emphasized the need for customized explanations that address the specific requirements of users across diverse scenarios. We plan to explore the application of CFEs in real-life scenarios including gender bias and recommendation systems, to showcase the effect of the different properties of the CFEs, and help us improve our analysis. 

Moving forward, it is imperative for researchers and practitioners in the XAI community to continue exploring the nuances of CFEs and developing methodologies for customizing explanations to meet the evolving needs of users. By incorporating user-centric design principles and considering diverse application contexts, we can further advance the field of XAI and empower users to engage more effectively with AI systems. Ultimately, our work contributes to the ongoing dialogue on CFEs and their role in shaping the future of transparent and trustworthy AI technologies. 

\section*{Acknowledgments}
This research work is Co-funded from the European Union’s Horizon Europe Research and Innovation programme under Grant Agreement No 101070631 and from the UK Research and Innovation(UKRI) under the UK government’s Horizon Europe funding guarantee (Grant No 10039436).- FSTP – Pilot Project:- SURE-GB

\balance{} 

\bibliographystyle{SIGCHI-Reference-Format}
\bibliography{extended-abstract}

\end{document}